\documentclass{article}

\usepackage{arxiv}

\usepackage[utf8]{inputenc} % allow utf-8 input
\usepackage[T1]{fontenc}    % use 8-bit T1 fonts
\usepackage{hyperref}       % hyperlinks
\usepackage{url}            % simple URL typesetting
\usepackage{booktabs}       % professional-quality tables
\usepackage{amsfonts}       % blackboard math symbols
\usepackage{nicefrac}       % compact symbols for 1/2, etc.
\usepackage{microtype}      % microtypography
\usepackage{lipsum}
\usepackage{array}
\usepackage{graphicx}
\usepackage{amsmath}
\usepackage{gensymb}

\title{A Machine Learning Framework for Real-time Inverse Modeling and Multi-objective Process Optimization of Composites for Active Manufacturing Control}

\author{
  Keith D. Humfeld \\
  Boeing Research and Technology\\
  Seattle, USA \\
  \texttt{Keith.D.Humfeld@boeing.com} \\
  \And
  Dawei Gu \\
  Department of Materials Sciences and Engineering\\
  University of Washington\\
  Seattle, USA \\
  \texttt{daweigu@uw.edu} \\
  \And
  Geoffrey A. Butler \\
  Boeing Research and Technology\\
  Seattle, USA \\
  \texttt{geoffrey.a.butler@boeing.com} \\
  \And
  Karl Nelson \\
  Boeing Research and Technology\\
  Seattle, USA \\
  \texttt{karlmarius@comcast.net} \\
  \And
  Navid Zobeiry \\
  Department of Materials Sciences and Engineering\\
  University of Washington\\
  Seattle, USA \\
  \texttt{navidz@uw.edu} \\
}

\begin{document}
\maketitle

\begin{abstract}
For manufacturing of aerospace composites, several parts may be processed simultaneously using convective heating in an autoclave. Due to uncertainties including tool placement, convective Boundary Conditions (BCs) vary in each run. As a result, temperature histories in some of the parts may not conform to process specifications due to under-curing or over-heating. Thermochemical analysis using Finite Element (FE) simulations are typically conducted prior to fabrication based on assumed range of BCs. This, however, introduces unnecessary constraints on the design. To monitor the process, thermocouples (TCs) are placed under tools near critical locations. The TC data may be used to back-calculate BCs using trial-and-error FE analysis. However, since the inverse heat transfer problem is ill-posed, many solutions are obtained for given TC data. In this study, a novel machine learning (ML) framework is presented capable of optimizing air temperature cycle in real-time based on TC data from multiple parts, for active control of manufacturing. The framework consists of two recurrent Neural Networks (NN) for inverse modeling of the ill-posed curing problem at the speed of 300 simulations/second, and a classification NN for multi-objective optimization of the air temperature at the speed of 35,000 simulations/second. A virtual demonstration of the framework for process optimization of three composite parts with data from three TCs is presented. 
\end{abstract}

% keywords can be removed

\section{Introduction}
For processing of aerospace composites, parts are commonly heated via convection in ovens or autoclaves. To ensure end-part quality, part temperature and pressure histories must conform to specifications (i.e. specs) which may be defined based on maximum part temperature or part temperature rates at transitions (e.g. transition from heat-up to dwell) [1,2]. Out-of-spec parts may suffer from a variety of process-induced defects such as under-curing, overheating, porosity, fiber waviness/wrinkling, and high degree of residual stresses leading to dimensional changes, micro-cracks and reduced mechanical performance [3–10]. A combination of experimental thermal profiling and numerical process simulation are used to design robust cure cycles conforming to all specs [11–18]. Based on these studies, lagging and leading locations (i.e. hottest and coldest zones) of parts are identified. Typically, thermocouples (TCs) are placed near these locations at the backside of the tool as proxies to monitor part temperature history. For large aerospace parts (e.g. fuselage or wing skin) where a dedicated autoclave is used, the distribution of convective BCs around the part can be measured with a relatively high degree of confidence in advance of manufacturing [12,19–21]. These BCs are then used as inputs to conduct process simulation using commercial Finite Element (FE) packages (e.g. [22]). Through an iterative and trial-and-error approach, process parameters including air temperature profile in the autoclave (i.e. cure cycle) are optimized to ensure that every point in the part satisfies process specifications. Based on this exhaustive search, a robust and optimized cure cycle is designed to minimize risk in manufacturing. To speed-up this process, reduced order FE modeling is commonly used  [12,23]. Away from the part edges and tooling sub-structures, the most dominant mode of heat transfer for thin composite parts is through-thickness. Consequently, 1D FE simulation of the curing process is frequently to speed-up analysis in the initial design phase [1,2,13]. Detailed designs are then conducted with limited number of 3D FE simulations [24].

\begin{figure}
  \centering
  \includegraphics{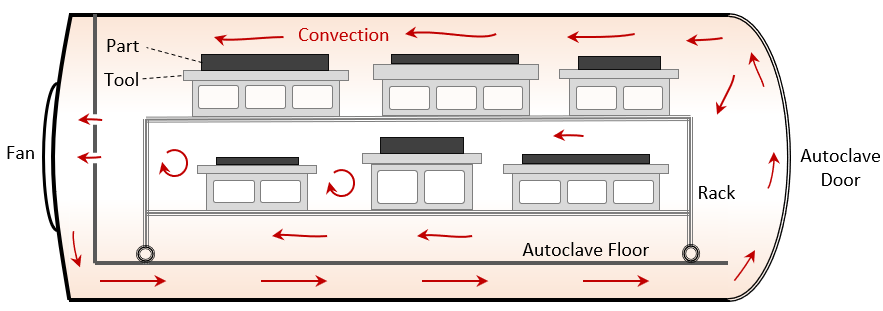}
  \caption{Schematic of the cross section of an autoclave for convective heating and curing of multiple parts simultaneously.}
  \label{fig:fig1}
\end{figure}

However, for cases where multiple parts are cured together in an autoclave/oven, convective BCs are typically not known before processing. This is due to many factors that affect the airflow pattern and velocity including the number of parts cured together, tool nesting and orientation, tool and part geometries and overall thermal mass of the load [20,25]. Figure \ref{fig:fig1}, shows a schematic of cross section of a floor air duct autoclave where 6 parts on tools are placed on two rows of a loading rack and cured simultaneously. Depending on the tool placement on the rack (e.g. distance to other tools, or distance to autoclave door), Heat Transfer Coefficients (HTCs) around parts and tools vary significantly [20]. In some cases, the airflow may be blocked reaching a part surrounded by other parts. As a result, this part may experience very low HTCs compared to other parts. Considering parts and tools with different geometries, this results in different temperature histories in the parts. Consequently, some of these parts may go out of spec due to under-curing or over-heating. 

To take advantage of analysis tools and mitigate manufacturing risk, this case can be decoupled into two problems: 1) the BCs around all parts and tools must be determined in each autoclave run, and 2) based on these BCs, an optimal cure cycle must be designed to satisfy specs in all parts. The main issue is the uncertainty of autoclave loading which results in uncertainty of BCs. The current industrial approach is to consider a wide range of viable BCs in an autoclave. Based on this initial assumption, a conservative cure cycle is designed using trial-and-error FE analysis to satisfy specs in different loading configurations. However, this introduces unnecessary constraints on the solution to increase processing time and reduce the production rate. The initial assumption for BCs, may also introduce additional risk in the solution. In another approach, data from tool TC may be used to back-calculate BCs using trial-and-error FE simulations. Given that this is a time-consuming process, even with reduced order 1D FE modeling, it cannot be implemented for real-time modeling and optimization of the cure cycle during processing. 
In recent years, significant progress has been achieved in implementing advanced analytics and Machine Learning (ML) approaches in various engineering applications including process simulation and failure analysis of composites [24,26–33]. One of the major trend is developing surrogate ML models based on FE results to significantly speed-up analysis compared to original FE model [24,26,27,29]. For example, in a recent study to speed-up thermo-chemical analysis of large composite parts, simulation speed was increased by 1000 times using a surrogate ML model compared to reduced order FE modeling [24]. By significantly speeding up analysis, real-time modeling and optimization tools can be developed based on manufacturing data [27]. Following these advances, for the current manufacturing problem, a surrogate ML model may be trained to use TC data as input to predict unknown BCs. The results from this model can be then used by a second ML model for fast optimization of the cure cycle. There are however complexities with such an approach. In a forward problem, FE uses known BCs as inputs to predict part and tool temperature histories. In the inverse problem, the ML model must use temperature histories from TCs as inputs, to predict BCs. Unfortunately, the inverse heat transfer problem, and to extension the inverse thermo-chemical problem of composites processing, is an ill-posed problem [27,34,35]. This means that for a given temperature profile, there are multiple BC solutions to yield similar responses. As a result, a robust ML model must be able to obtain all potential BC solutions based on TC data. In addition, the cure cycle optimization is a multi-objective problem. This requires developing extremely fast ML models to iterate a large number of potential solutions in real-time and obtain an optimal design.

In this paper, a recently developed ML framework is presented, capable of solving the ill-posed inverse thermo-chemical problem during composites manufacturing in real-time to obtain all potential BC solutions. The ML framework then uses all potential solutions to conduct a multi-objective optimization to obtain the shortest cure cycle that can satisfy manufacturing specs in all parts in an autoclave. The optimization tool is capable of conducting more than 35,000 simulations per second. This allows for real-time optimization of cure cycle for active manufacturing control while processing multiple composite parts. A successful demonstration is presented where the air temperature profile for curing three HEXCEL AS4/8552 parts on three Invar tools is optimized based on data from three TCs. 

\section{Background}

Consider a 1D representation of a composite part with a thickness of L1, placed on a tool with a thickness of L2, and cured subjected to an air temperature profile of \(T_{air}(t)\) and heat transfer BCs, h1 and h2, as shown in Figure 2a. Two thermocouples (TC) are used to measure temperatures at the center of the part, \(T_{part}(t)\), and tool back-side, \(T_{tool}(t)\). In practice, no TC is usually placed in the part. Instead, several TCs are placed at the tool back-side (i.e. proxy TCs). Assume that the part and tool are subjected to a two-hold air temperature profile which consists of three heating and cooling ramps (\(T_1^\prime\), \(T_2^\prime\),\(T_3^\prime\) ), two dwell temperatures (\(T_1,T_2\)), and two dwell times (\(t_1,t_2\)) as shown in Figure 2b. During the heat-up, part and tool temperatures initially lag behind the air temperature (Figure 2b). This is due to the tool and part thermal masses, as well as combined convective and conductive thermal resistances [19,20]. Once the exothermic curing reaction starts in the part, \(T_{part}\) starts increasing beyond the air temperature to reach a maximum temperature of \(T_{max}\). During this step, a metallic tool typically acts as a heat sink to demonstrate a lower temperature profile. The difference in part and temperature profiles is further increased in practice given that the convective BC under the tool is typically lower than the BC above the part (i.e. h1 > h2). Usually process specifications limit maximum part temperature (\(T_{max}\)), as well as part temperature rates in transition zones (e.g. \(T_{part}^\prime\) in Figure 2b). To predict the temperature distribution in the part and check for conformity to process specifications, thermo-chemical analysis may be performed by including the effect of heat generation in the composite part. For the 1D example in Figure 2a, the system of governing Partial Differential Equations (PDEs) are listed here:

\begin{equation}
\begin{aligned}
& h_1\ (T_{air}-T_1\ )=k_1 (\frac{\partial T_1}{\partial z}) \ \ \ \ \ \ \ \ \ \ \ \ \ \ \ \ \ \ \ \ \ z=L_1 \\
& \frac{\partial}{\partial t}(\rho_1\ C_{P1}\ T_1\ )=\frac{\partial}{\partial z}(k_1 \frac{\partial T_1}{\partial z})+\dot{Q} \ \ \ \ \ \ z\in[0,L_1]\\
& k_1 \frac{\partial T_1}{\partial z}=k_2 \frac{\partial T_2}{\partial z} \ \ and \ \ T_1=T_2 \ \ \ \ \ \ \ \ \ \ \ \ \ z=0  \\      
& \frac{\partial}{\partial t}(\rho_2\ C_{P2}\ T_2\ )=\frac{\partial}{\partial z}(k_2 \frac{\partial T_2}{\partial z}) \ \ \ \ \ \ \ \ \ \ \ \ \ \ z\in[-L_2,0] \\
& h_2 (T_2-T_{air} )=k_2 (\frac{\partial T_2}{\partial z}) \ \ \ \ \ \ \ \ \ \ \ \ \ \ \ \ \ \ \ \ \ \  z=-L_2
\end{aligned}
\end{equation}

\begin{figure}
  \centering
  \includegraphics[width=5.9in]{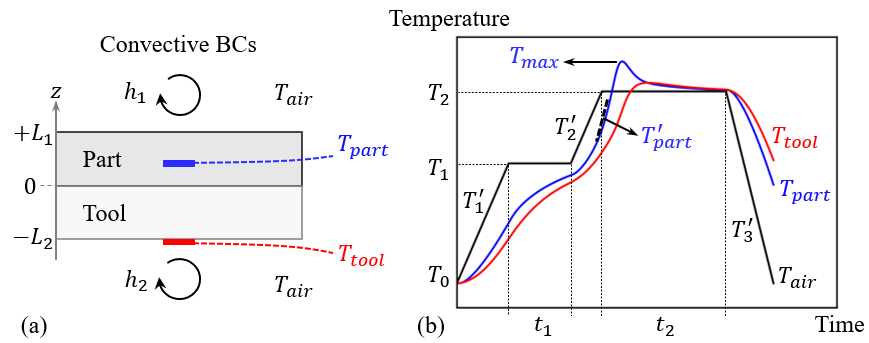}
  \caption{(a) 1D representation of a part cured on a tool with convective heating, and (b) time-temperature history of tool and part thermocouples (TCs) subjected to a two-hold air temperature cycle.}
  \label{fig:fig2}
\end{figure}

In these equations, subscript 1 refers to the part, and subscript 2 refers to the tool. Here, \(\rho\) is the density, \(C_P\) is the specific heat capacity, \(k\) is the thermal conductivity, and \(\dot{Q}\) is the rate of heat generation in the composite part due to the exothermic polymerization reaction. Z direction is the through-thickness direction as shown in Figure 2a. The first and last PDEs are convective heat transfers at the two boundaries, the second and forth PDEs are conductive heat transfers in the part and tool, and the third PDE is the conductive BC between tool and part. This is considered a forward problem where material properties, geometries, BCs, and initial conditions (ICs) are used as inputs to calculate temperature histories. We can solve this forward problem using FE analysis to predict unique solutions for part and temperature histories.

Instead of FE analysis, we can conduct an experiment where a TC is placed under the tool to measure the tool temperature history. In this case, however, results are affected by measurements noise and errors:

\begin{equation}
\begin{aligned}
T_{tool}=f_1(T_{air},L_1,L_2,h_1,h_2)+noise
\end{aligned}
\end{equation}

Assuming that convective BCs are not known, by measuring tool temperature, we can back-calculate BCs in an inverse problem: 

\begin{equation}
\begin{aligned}
h_1,h_2=f_2 (T_{air},T_{tool},L_1,L_2,noise)
\end{aligned}
\end{equation}

As described previously, this is an ill-posed problem. There are two main issues with solving this inverse equation. For a given tool temperature profile, there are multiple solutions for BCs. In addition, these solutions are highly affected by the level of noise in measurements. The ill-posedness of this problem is demonstrated with an example in Figure 3. In this example, a 20 mm composite part (HEXCEL AS4/8552 [36]) is cured on a 15 mm Invar tool. Subjected to a two-hold air temperature cycle, and BCs of \(h_1=60 \ and \ h_2=40 W/m^2 K\),  a baseline tool temperature, \(T_a\), was predicted using FE analysis as shown in Figure 3a (details of FE analysis will be discussed in the next section). After this, many FE simulations were conducted with different values for BCs. For each simulation, the maximum error between predicted tool temperature, \(T_i\), and baseline temperature were measured:

\begin{equation}
\begin{aligned}
Error=Max|T_i(t)-T_0(t)|
\end{aligned}
\end{equation}

Contours of these error values for different BCs are plotted in Figure 3b. This shows that for many combinations of BCs, negligible differences in temperature profiles are predicted. Two specific cases, \(T_b\) and \(T_c\), are compared with baseline, \(T_a\) in Figure 3a and listed below:

\begin{itemize}
\item \(T_a: h_1=60 \ and \ h_2=40 \ W/m^2 K\)
\item \(T_b: h_1=46.7 \ and \ h_2=42.6 \ W/m^2 K, \ Error = 1.0 \ \degree C\)
\item \(T_c: h_1=79.5 \ and \ h_2=38.5 \ W/m^2 K, \ Error = 0.95 \ \degree C\)
\end{itemize}

In fact, there are many acceptable combinations of BCs with maximum errors less than 1 \degree C. This is close to ±1.1 \degree C which is the Special Limits of Error (SLE) for a type J thermocouple. There are several approaches proposed in the literature to solve the ill-posed inverse problems including regularization [37] and utilizing prior information [38]. But most of these established methods cannot identify all potential solutions within a given tolerance. For manufacturing problems, we are interested in all potential solutions to mitigate risk. One viable approach is to assume a range of potential BCs, and then solve the forward problem with many combinations of BCs within these ranges. Using ML, such a process can be done quite fast and in near real-time. Upon obtaining all BCs (\(h_{1_i},h_{2_i}\)), the problem then can be solved in a forward manner to predict all potential part temperature histories:

\begin{equation}
\begin{aligned}
T_{part_i}=g(T_{air},L_1,L_2,h_{1_i},h_{2_i})
\end{aligned}
\end{equation}

From part temperature profiles, conformity to process specs in all solutions can be verified. Based on this, we can conduct a multi-objective optimization to minimize the total cycle time:

\begin{itemize}
\item Minimize the total cycle time:
    \begin{equation}
    \begin{aligned}
    min \sum_{i=1}^{m} \frac{T_i-T_{i-1}}{\dot{T_i}}+t_i
    \end{aligned}
    \end{equation}
\item Subject to satisfying all process specifications in all parts (example):
    \begin{equation}
    \begin{aligned}
    &T_{max_i} < limit_1 \ \ \ \ i\in[1,n] \\
    &limit_2 < \dot{T_{part_i}} < limit_3 \ \ \ \ i\in[1,n]
    \end{aligned}
    \end{equation}
\end{itemize}

Where \(m\) is the number of steps in the air temperature profile (ramps followed by dwell) as shown in Figure 2, and n is the total number of plausible BC solutions for multiple parts in an autoclave. In addition to above specifications, others may be required. For example, typically the final hold temperature and time (example, \(T_2\)  and \(t_2\) in Figure 2) are specified to achieve a certain level of degree of cure by the end of the cycle. Additional limitations are also imposed on the cool-down rate. 

\begin{figure}
  \centering
  \includegraphics[width=5.9in]{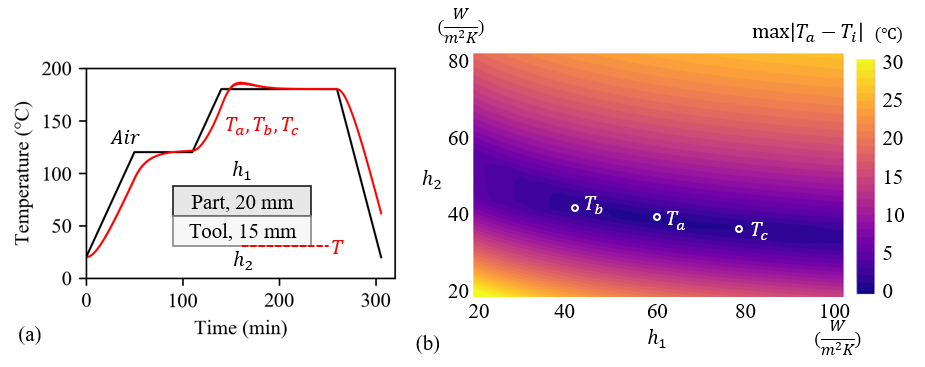}
  \caption{Demonstration of ill-posedness of the thermo-chemical problem during composites processing: (a) three similar time-temperature histories for a 20 mm HEXCEL AS4/8552 composite part on a 15 mm Invar tool subjected to different BCs, and (b) error contour of different BCs compared to baseline \(T_a\).}
  \label{fig:fig3}
\end{figure}

\section{A Machine Learning Framework}
\subsection{Architecture }
To develop a machine learning framework, the problem described in previous section is decoupled into two problems: 1) inverse modeling of the thermo-chemical curing problem based on tool TC data to identify all potential BCs within an acceptable tolerance, and 2) multi-objective optimization of the air temperature profile to identify the shortest cycle to satisfy all process specifications in all potential solutions. For each of these, a ML model was developed to form a framework for real-time analysis and optimization of the manufacturing problem as schematically shown in Figure 4. 
Considering the ill-posedness of the inverse governing equations for convective heating and curing of composites, a forward ML framework was developed for fast evaluation of a wide range of BCs. By comparing TC data with predictions of different BCs, all plausible solutions can be identified (i.e. maximum error < 1 \degree C).  For this purpose, two Neural Networks (NNs) with LSTM (Long short-term memory) architectures [39] were designed to predict part and tool temperatures for a given thermal stack (\([h_1,L_1,L_2,h_2]\)) and air temperature profile (Figure 5).  The main advantage of these models over FE is the speed. On a typical computer workstation, these trained LSTM models can perform about 300 1D thermo-chemical simulations of composites per second. This allows the LSTM models to quickly scan a wide range of BCs and identify cases that result in similar temperature histories to measured tool TC data. 
A third NN was designed with a simple feedforward architecture (Figure 6) for multi-objective optimization of the temperature cycle. For a given thermal stack, this NN can classify if an air temperature profile satisfies all manufacturing specs or not. On a typical workstation, this NN can classify about 35,000 cases per second. This enables scanning a wide range of air temperature profiles and identify the shortest cycle to satisfy process specs in all parts. 

Using this framework and based on the data continuously gathered from proxy TCs, all BC solutions for all parts in an autoclave load can be identified. Based on these solutions, the cure cycle can be optimized in real-time for active manufacturing control as schematically shown in Figure 4. 

\begin{figure}
  \centering
  \includegraphics[width=5in]{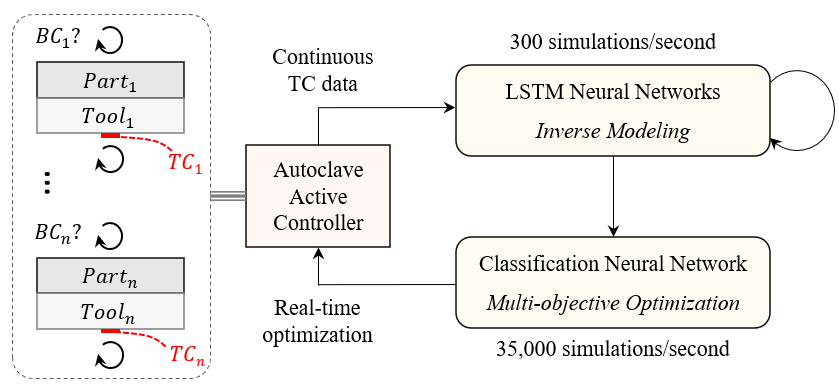}
  \caption{Schematic of the ML framework for inverse modeling of the composites processing, and optimization of the air temperature profile in real time for active manufacturing control.}
  \label{fig:fig4}
\end{figure}

\begin{figure}
  \centering
  \includegraphics[width=5in]{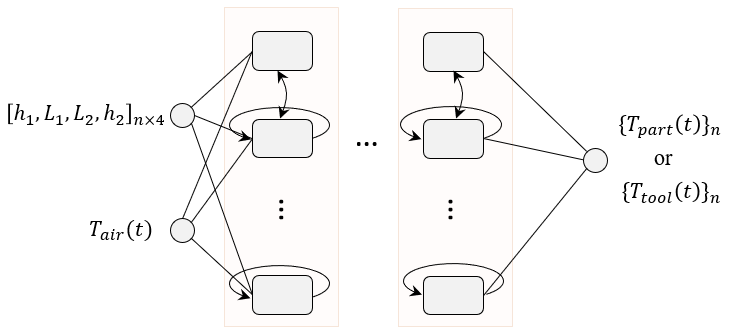}
  \caption{Two LSTM Neural Networks to predict time-temperature histories of multiple parts and tools.}
  \label{fig:fig5}
\end{figure}

\begin{figure}
  \centering
  \includegraphics[width=4.6in]{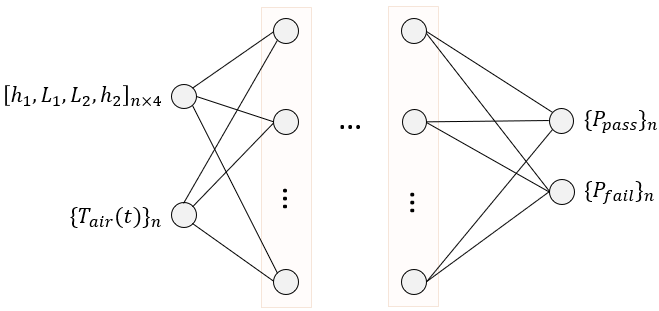}
  \caption{a Feed Forward Neural Network to classify thermal stacks and temperature profiles to satisfy manufacturing specifications.}
  \label{fig:fig6}
\end{figure}

\subsection{Training Data}
ML training data was generated using an in-house developed and python-based FE code for 1D thermo-chemical analysis of composites processing. A schematic of the 1D problem is shown in Figure 2a where a composite part is cured via convection on a tool.  For the composite material, HEXCEL AS4/8552 [40], and for the tooling material, Invar-36 were used in all simulations. Thermo-chemical properties of AS4/8552 prepreg system, as well as thermal properties of Invar-36 have been extensively studied and reported in the literature [5,8,11,12,36,41,42]. This includes cure kinetics model of 8552 epoxy resin system, as well as density, thermal conductivity and specific heat capacity of AS4/8552 prepreg system and Invar-36 [12,36]. These properties were implemented as inputs to solve the 1D problem using FE by discretizing tool and part to 10 elements each. As an example, FE simulation results for part-center and tool-bottom temperatures subjected to a specific air temperature are shown in Figure 2b. The in-house FE model was validated against a commercially available FE tool, RAVEN [43], with similar material properties. The developed FE model was then used to conduct 100,000 simulations by randomly varying 8 FE input parameters. 4 parameters were varied to define a complete thermal stack as shown in Figure 2a: 

\begin{itemize}
    \item \(20<h_1 \ and \ h_2 <100 \ W/m^2 K\)
    \item \(2<L_1<20 \ mm\)
    \item \(8<L_2<20 \ mm\)
\end{itemize}
	
In addition, 4 parameters were varied to define a two-hold air temperature cycle (i.e. cure cycle) as shown in Figure 2b:

\begin{itemize}
    \item \(1<\dot{T_1} \ and \ \dot{T_2}<8 \ \degree C/min\)
    \item \(110<T_1<180 \ \degree C\)
    \item \(0<t_1<120 \ min\)
\end{itemize}
	
The other parameters of the cure cycle were kept constant for data generation: \(T_0=20 \ \degree C, T_2=180 \ \degree C, t=120 \ 
 \degree C/min\), followed by a final cool-down rate of 3.5 \degree C/min to room temperature of 20  On average for a given 1D case, simulation time was less than one second on a typical computer work station (about 24 hours for 100,000 simulations on an Intel i9 CPU with 32 GB RAM and no GPUs). From each simulation, temperature profile at the center of the part, as well as bottom of the tool were recorded. In addition, the maximum part temperature in the cycle (\(T_{max}\) in Figure 2b), and the part temperature rate when air temperature reaches 180 \degree C  (\(\dot{T_{part}}\) in Figure 2b), were recorded in each simulation.

\subsection{Inverse Modeling using LSTM Neural Networks }
Part and tool time-temperature profiles from 100,000 FE simulations were used to train surrogate machine learning models. These models take 8 parameters as inputs (4 for the thermal stack and 4 for the air temperature profile as described in the previous section) and predict part and tool temperature histories. This is similar to FE but significantly faster. For this purpose, two Recurrent Neural Networks with Long Short-Term Memory (LSTM) architectures were trained as schematically shown in Figure 5. One of these LSTM models predicts temperature at the bottom of the tool, \(T_{tool}(t)\), and the other one temperature at the center of the part \(T_{part}(t)\). Training of the models were done using Tensorflow API (version 2.1) [44] available in Python (version 3.6.8). FE results were divided into 70\% training data and 30\% validation data. Using the grid search, optimal hyper-parameters for training LSTM models were obtained as reported in Table 1. After 1000 iterations, LSTM prediction errors compared to FE results were reduced less than 1 ℃. For validation, comparison of LSTM predictions and FE results for a specific case is shown in Figure 7:

\begin{itemize}
    \item \([h_1,L_1,L_2,h_2 ] = [40 \ W/m^2K, 15 \ mm, 12 \ mm, 24 \ W/m^2K]\)
    \item \([\dot{T_1},\dot{T_2},T_1,t1]=[2 \ \degree C/min,2 \ \degree C/min,120 \ \degree C,60 \ min]\)   
\end{itemize}

LSTM maximum error compared to FE was equal to 0.73 ℃. To further speedup simulations, LSTM architecture was designed to take multiple thermal stacks as inputs, \([h_1,L_1,L_2,h_2]_{n\times 4}\), along with one specific cure cycle, \(T_{air}(t)\), and predict part and tool temperature histories for these cases simultaneously (\({T_{part}(t)}_n,{T_{tool}(t)}_n\)). Using this approach, trained LSTM models conduct more than 300 1D simulations per second on a typical computer workstation. This is a speed improvement of about 300 times compared to the original FE model in python. This is also a speed improvement of about 1000 times compared to the commercial FE tool used for validation.

\begin{table}[]
    \centering
        \includegraphics[width=5.9in]{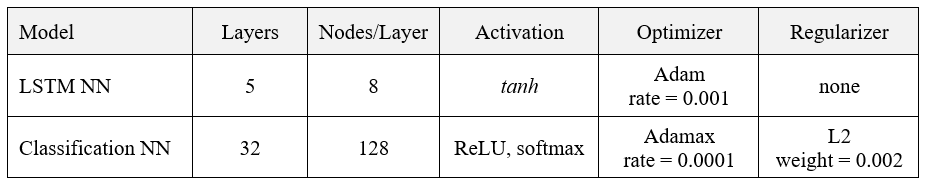}
    \caption{Optimal hyper-parameters used for training machine learning models.}
    \label{tab:table1}
\end{table}

\begin{figure}
  \centering
  \includegraphics[width=4in]{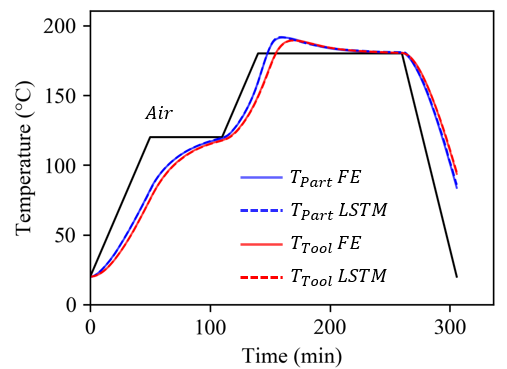}
  \caption{Comparison of the capability of the LSTM NN and the FE model to predict part and tool temperature histories in a 15 mm part (HEXCEL AS4/8552) cured on a 12 mm tool (Invar) and subjected to a two-hold temperature cycle with BCs of 40 and 24 \(W/m^2K\). Both predict identical solutions. }
  \label{fig:fig7}
\end{figure}

\subsection{Multi-objective Optimization using a Classification Neural Network}
For multi-objective optimization of processing parameters, a classification Neural Network with a feedforward architecture as shown in Figure 6 was trained. Similar to previous ML models, 100,000 FE simulations were used for this training. From each FE simulation, two parameters were recorded: maximum part temperature, \(T_{max}\), and part temperature rate, \(\dot{T_{part}}\), as shown in Figure 2b. These values were then compared to the following process specifications to determine if a cycle is acceptable:

\begin{equation}
\begin{aligned}
T_{max}<185 \ \degree C \ \ and \ \ 1 \degree C<\dot{T_{part}}<3 \ \degree C/min
\end{aligned}
\end{equation}

All cycles were then classified as pass or fail (score of one or zero) by conducting the above comparison. For training of the classification NN model, thermal stack parameters \([h_1,L_1,L_2,h_2]_{n\times 4}\) and air temperature parameters (\(\{T_{air}(t)\}_n\)) were used as inputs. But instead of predicting part and tool temperatures, classification model was trained to predict the failure and pass probabilities (\(\{P_{pass}\}_n,\{P_{fail}\}_n \)).  Steps described in previous section were followed for training. Hyper-parameters of the trained model are listed under Table 1. To speed-up machine learning simulations, classification model was designed to take multiple thermal stacks and thermal cycles as inputs. The final trained model conducts more than 35,000 simulations per second to identity cycles that satisfy process specifications. Such a speed enables real-time optimization of process parameters. 

\section{Implementation and Validation}
The framework described in previous section for data generation, ML training, and TC data evaluation for process optimization, was implemented in a python-based machine learning software developed at the University of Washington, CompML (Composites Machine Learning) [27,29]. For demonstration, CompML is used here for inverse modeling and optimization of cure cycle for simultaneous processing of three parts. We consider curing of three HEXCEL AS4/8552 composite parts with different thicknesses on three Invar tools as schematically shown in Figure 8. 1D thermal stacks for these parts are listed below:

\begin{itemize}
    \item 1: \(h_1=60 \ W/(m^2K),L_1=10 \ mm,L_2=10 \ mm,h_2=40 \ W/m^2K\)
    \item 2: \(h_1=40 \ W/(m^2K),L_1=15 \ mm,L_2=10 \ mm,h_2=40 \ W/m^2K\)
    \item 3: \(h_1=80 \ W/(m^2K),L_1=20 \ mm,L_2=10 \ mm,h_2=40 \ W/m^2K\)
\end{itemize}

\begin{figure}
  \centering
  \includegraphics[width=5.9in]{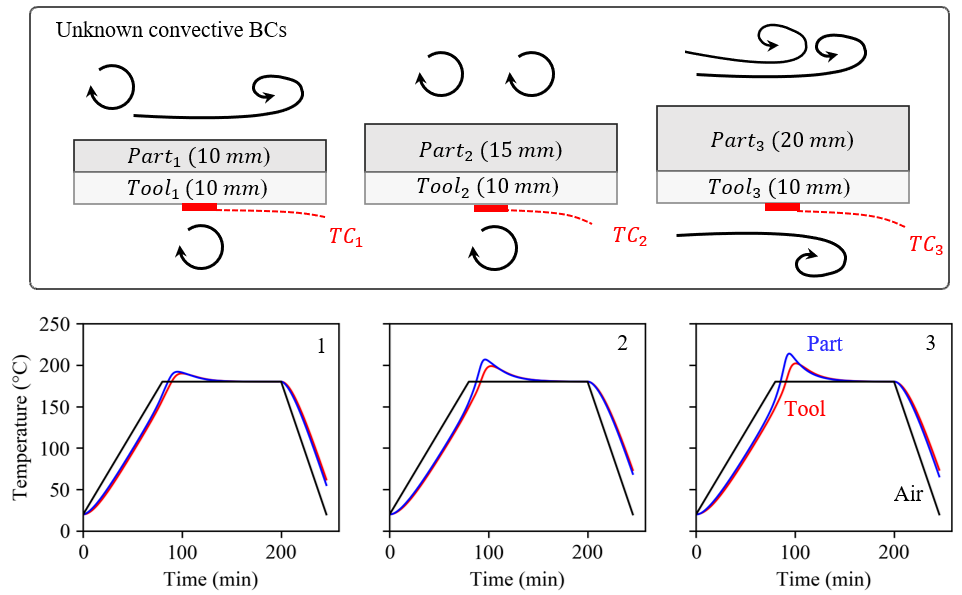}
  \caption{Schematic of processing of three parts on three tools with unknown convective BCs while subjected to a one-hold temperature cycle. For a case with known BCs, air temperature, tool back-side temperature, and part center temperature are predicted using FE analysis.}
  \label{fig:fig8}
\end{figure}

For this example, two specifications for the part maximum temperature and part temperature rate were considered:

\begin{itemize}
    \item \(T_{max}<185 \ \degree C\)
    \item \(1 \degree C<\dot{T_{part}(@T_{air}=180 \ \degree C)}<3 \ \degree C/min\)
\end{itemize}

The three parts are subjected to a one-hold cycle from room temperature to 180 ℃ with a heating rate of 2 ℃/min, followed by two hours hold and then cool-down to room temperature with a rate of 3.5 ℃/min. For the given thermal stacks and known BCs, part and tool temperature histories were predicted using FE as shown in Figure 8. Based on this analysis, none of the parts satisfy the specifications. 
For application of the ML framework, we assumed unknown BCs. Following steps were then taken by CompML software:

\begin{itemize}
    \item For unknown BCs, values between 20-100 W/m2K were considered for evaluation with LSTM models with a step size of 5 W/m2K (i.e. 16 values). Consequently, for each part with 2 unknown BCs, 256 (16×16) potential thermal stacks were considered. For 3 parts, a total of 768 cases were considered (16×16×3). 
	\item Initially, we assumed that 15 minutes of the start of cycle has passed (i.e. 15 minutes of TC data). CompML predicted the tool temperatures for all 768 thermal stacks and compared with tool TC data from the first 15 minutes. The entire calculation took 2.5 seconds to complete on a typical workstation. From evaluations of these 768 cases, CompML selected 138 plausible BC solutions based on similarity to tool TC data with a maximum error of 1 ℃. CompML then predicted the entire part temperature histories for these 138 cases and plotted them to create zones of plausible solutions as shown in Figure 9. For validation, these zones were compared with the FE predictions with known BCs (i.e. hidden true responses). This comparison validates the capability of CompML to identify correct zones of solutions with unknown BCs.
    \item For the second step, TC data from the first 30 minutes of the cycle was used. However, instead of starting from the original 768 cases, CompML started with the identified 138 thermal stacks from the first 15 minutes. In a similar calculation to previous step, CompML reduced these to 58 potential thermal stacks in 0.9 second. Out of these 58 cases, 21 were plausible BC solutions for the first part, 18 for the second part, and 19 for the third part. The zones of predictions are shown in Figure 10 and successfully compared with FE predictions. This validates the capability of CompML to correctly identify the zones of potential solutions based on analysis results from the first 15 minutes, and TC data from the first 30 minutes.
    \item 58 potential solutions were then transferred to the classification NN model to optimize the cure cycle. By varying \(T_1\) between 110 and 180 ℃ with a step size of 5 \degree C, \(t_1\) between 0 and 120 minutes with a step size of 5 minutes, and \(\dot{T_2}\) between 1 to 8 ℃/min and a step size of 0.2 ℃/min, total of 11760 cycles were constructed for evaluation. CompML went through the 58 potential thermal stacks one at a time and evaluated conformity to process specifications with selected thermal cycles. After evaluating each thermal stack, rejected cycles were removed from the original number of cycles, and evaluation continued to the next thermal stack. Through this process of elimination of cycles, total of 81418 cases (i.e. different combinations of thermal stack and cycles) were evaluated in only 2.3 seconds. From these evaluations, 280 cycles were selected that could satisfy all process specifications for all 58 potential BC solutions. 
    \item Identified 280 cure cycles from previous step were sorted based on the total cycle time, and the shortest cycle was selected. Using LSTM models, the part temperature histories in all 58 cases subjected to the optimal cure cycle were predicted and compared with FE predictions in Figure 11. 
\end{itemize}

\begin{figure}
  \centering
  \includegraphics[width=5.9in]{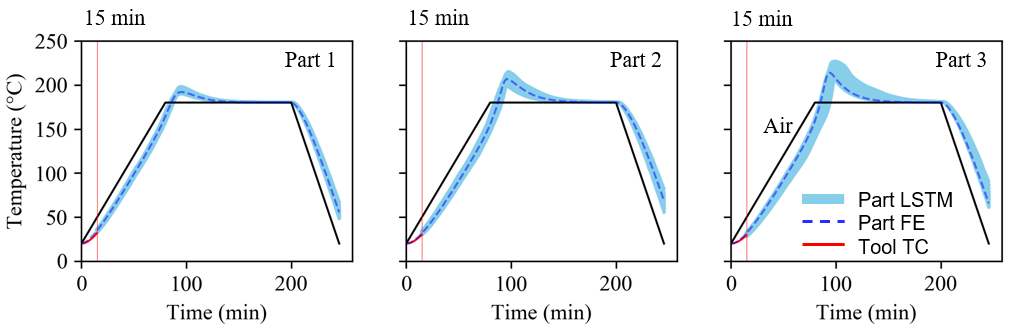}
  \caption{Based on the initial 15 minutes data from tool TCs, CompML predicts all solutions for part temperature histories. The zones of viable solutions are compared with the true repose predicted by FE.}
  \label{fig:fig9}
\end{figure}

\begin{figure}
  \centering
  \includegraphics[width=5.9in]{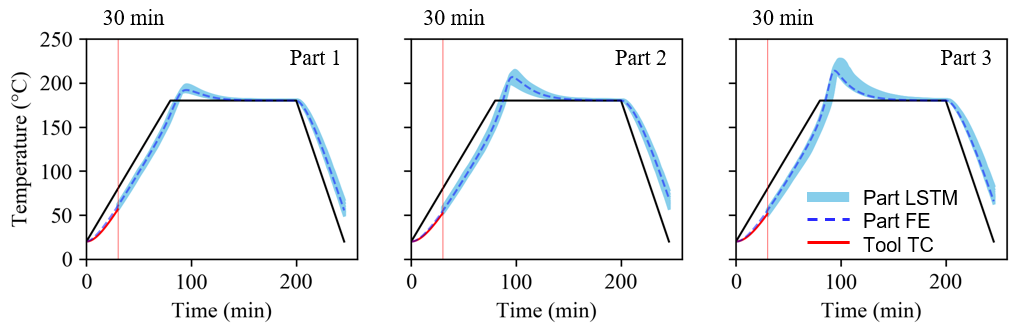}
  \caption{Based on the initial 30 minutes data from tool TCs, CompML predicts all solutions for part temperature histories. The zones of viable solutions are compared with the true repose predicted by FE.}
  \label{fig:fig10}
\end{figure}

\begin{figure}
  \centering
  \includegraphics[width=5.9in]{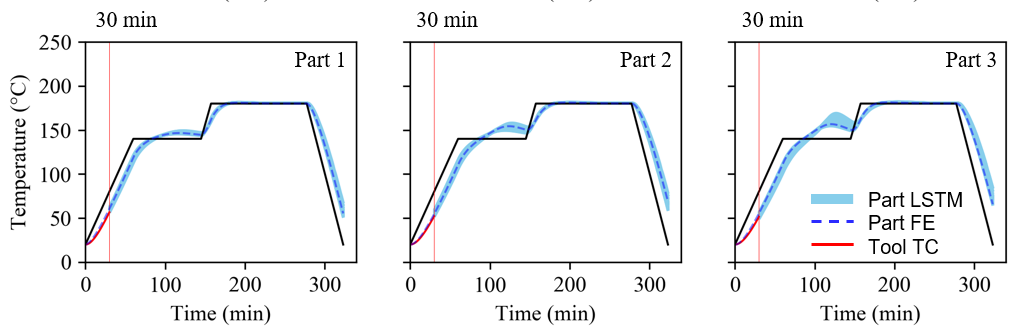}
  \caption{Based on potential BC solutions using the initial 30 minutes TC data, CompML optimizes the original cycle to a 2-hold cycle to satisfy specs in all potential solutions. The zones of viable solutions are compared the true repose predicted by FE.}
  \label{fig:fig11}
\end{figure}

While the maximum part temperature for all 58 cases is less than 181.4 ℃, the part temperature rate varies between 1.05 and 2.33 ℃/min. This satisfies process specifications. The parameters of the optimized two-hold cure cycle is compared with the original one-hold cycle in Table 2. True part temperature and part temperature rates in the original and optimized cycles are listed in Table 3. It should be noted that the original cycle could not satisfy process specifications in any of the parts due to excessive maximum part temperatures and temperature rates. While the original one-hold cycle was 245 minutes, the optimized cycle time is increased to 323 minutes to conform to specifications. It should be noted that the optimal cycle is obtained based on the shape of a two-hold cycle. If more complex cycle shapes are considered, or if smaller steps are selected for discretization of cycle parameters, the cycle time may be further reduced. The framework described here can be easily extended to accommodate these assumptions. 

\begin{table}[]
    \centering
        \includegraphics[width=5.9in]{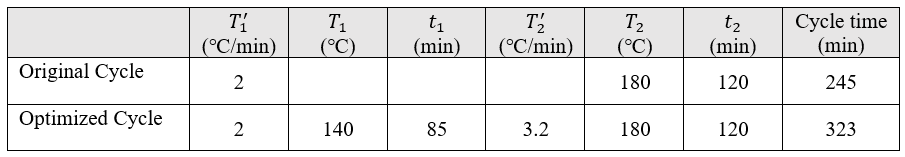}
    \caption{Parameters of the original one-hold cycle and optimized two-hold cycle based on the definitions in Figure 2.}
    \label{tab:table2}
\end{table}

\begin{table}[]
    \centering
        \includegraphics[width=5.9in]{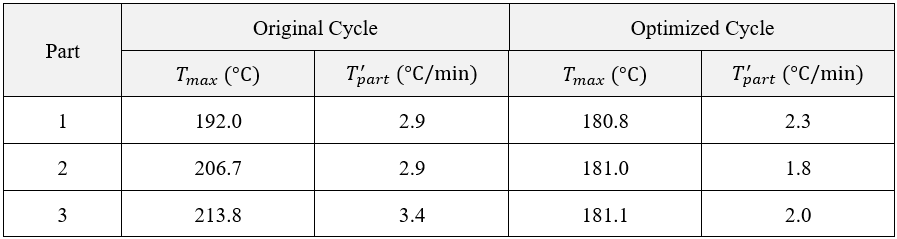}
    \caption{Maximum part temperatures, and part heating rates for the original and optimized cure cycles.}
    \label{tab:table3}
\end{table}

\section{Summary and Conclusions}
For fabrication of aerospace composite components, several parts may be processed together in a convective oven/autoclave. This introduces uncertainties in manufacturing, and results in variation of convective BCs in each autoclave load. In The current industrial approach to mitigate risk is based on conducting FE analysis of the thermo-chemical problem with an assumed range of BCs. This places unnecessary constraints on the final design, to increase the cycle time. In some cases, based on the initial assumption of BCs, manufacturing risk may be further increased. In this study, we introduced a novel ML framework for real-time optimization of composites processing based on tool TC data during manufacturing. The framework consists of several NNs for inverse modeling of the ill-posed thermo-chemical problem during processing, as well as multi-objective optimization of the air temperature profile to satisfy process specifications in multiple parts simultaneously. The framework was implemented in a python-based ML software, CompML.  For multiple parts heated and cured via convection, the capability of the framework was demonstrated to optimize the cure cycle based on tool TC data only, without the knowledge of convective BCs. Based on TC data, LSTM NN models were able to solve the forward problem to identify all plausible BC solutions in few seconds. The results were then used with a fast classification NN to identify the optimal air temperature cycle to satisfy specs in all parts. While this mitigates manufacturing risk associated with unknown BCs, the framework can be implemented for real-time optimization of curing process with active controlling of an oven/autoclave. The framework can be extended in the future to include more complex temperature cycles and different part and tooling materials. 

\section*{Acknowledgements}
We would like to acknowledge financial supports by the Boeing company, University of Washington, and State of Washington in USA.  

\section*{References}
\begin{enumerate}

\item Fabris J, Lussier D, Zobeiry N, Mobuchon C, Poursartip A. Development of standardized approaches to thermal management in composites manufacturing. Int. SAMPE Tech. Conf., 2014.
\item Fabris J, Mobuchon C, Zobeiry N, Lussier D, Fernlund G, Poursartip A, et al. Introducing thermal history producibility assessment at conceptual design. Int. SAMPE Tech. Conf., vol. 2015- Janua, 2015.
\item Mohseni M, Zobeiry N, Fernlund G. Experimental and numerical study of coupled gas and resin transport and its effect on porosity. J Reinf Plast Compos 2019;38:1055–66. https://doi.org/10.1177/0731684419865783.
\item Farhang L, Mohseni M, Zobeiry N, Fernlund G. Experimental study of void evolution in partially impregnated prepregs. J Compos Mater 2020;54:1511–23. https://doi.org/10.1177/0021998319883934.
\item Zobeiry N, Poursartip A. The origins of residual stress and its evaluation in composite materials. In: Beaumont PWR, Soutis C, Hodzic A, editors. Struct. Integr. Durab. Adv. Compos. Innov. Model. Methods Intell. Des., Woodhead Publishing; 2015, p. 43–72. https://doi.org/10.1016/B978-0-08-100137-0.00003-1.
\item Farnand K, Zobeiry N, Poursartip A, Fernlund G. Micro-level mechanisms of fiber waviness and wrinkling during hot drape forming of unidirectional prepreg composites. Compos Part A Appl Sci Manuf 2017;103:168–77. https://doi.org/10.1016/j.compositesa.2017.10.008.
\item Zobeiry N, Lee A, Mobuchon C. Fabrication of transparent advanced composites. Compos Sci Technol 2020;197:108281. https://doi.org/10.1016/j.compscitech.2020.108281.
\item Li C, Zobeiry N, Keil K, Chatterjee S, Poursartip A. Advances in the characterization of residual stress in composite structures. Int. SAMPE Tech. Conf., 2014.
\item Potter KD. Understanding the origins of defects and variability in composites manufacture. ICCM Int. Conf. Compos. Mater., 2009.
\item Potter K, Khan B, Wisnom M, Bell T, Stevens J. Variability, fibre waviness and misalignment in the determination of the properties of composite materials and structures. Compos Part A Appl Sci Manuf 2008;39. https://doi.org/10.1016/j.compositesa.2008.04.016.
\item Zobeiry N, Forghani A, Li C, Gordnian K, Thorpe R, Vaziri R, et al. Multiscale characterization and representation of composite materials during processing. Philos Trans R Soc A Math Phys Eng Sci 2016;374:20150278. https://doi.org/10.1098/rsta.2015.0278.
\item Zobeiry N, Park J, Poursartip A. An infrared thermography-based method for the evaluation of the thermal response of tooling for composites manufacturing. J Compos Mater 2019;53:1277–90. https://doi.org/10.1177/0021998318798444.
\item Fabris J, Mobuchon C, Zobeiry N, Poursartip A. Understanding the consequences of tooling design choices on thermal history in composites processing. Int. SAMPE Tech. Conf., vol. 2016- Janua, 2016.
\item Park J, Zobeiry N, Poursartip A. Tooling materials and their effect on surface thermal gradients. Int. SAMPE Tech. Conf., 2017, p. 2554–68.
\item Haghshenas SM, Vaziri R, Poursartip A. Integration of resin flow and stress development in process modelling of composites: Part I – Isotropic formulation. J Compos Mater 2018;52. https://doi.org/10.1177/0021998318762295.
\item Haghshenas SM, Vaziri R, Poursartip A. Integration of resin flow and stress development in process modelling of composites: Part II – Transversely isotropic formulation. J Compos Mater 2018;52. https://doi.org/10.1177/0021998318762296.
\item Niaki SA, Forghani A, Vaziri R, Poursartip A. An orthotropic integrated flow-stress model for process simulation of composite materials-part II: Three-phase systems. J Manuf Sci Eng Trans ASME 2019;141. https://doi.org/10.1115/1.4041862.
\item Niaki SA, Forghani A, Vaziri R, Poursartip A. An orthotropic integrated flow-stress model for process simulation of composite materials-part I: Two-phase systems. J Manuf Sci Eng Trans ASME 2019;141. https://doi.org/10.1115/1.4041861.
\item Hubert P, Fernlund G, Poursartip A. Autoclave processing for composites. Manuf. Tech. Polym. Matrix Compos., 2012. https://doi.org/10.1016/B978-0-85709-067-6.50013-4.
\item Fernlund G, Mobuchon C, Zobeiry N. 2.3 Autoclave Processing. In: Zweben C, Beaumont P, editors. Compr. Compos. Mater. II, vol. 2, Elsevier; 2018, p. 42–62. https://doi.org/10.1016/b978-0-12-803581-8.09899-4.
\item Bohne T, Frerich T, Jendrny J, Jürgens JP, Ploshikhin V. Simulation and validation of air flow and heat transfer in an autoclave process for definition of thermal boundary conditions during curing of composite parts. J Compos Mater 2018;52. https://doi.org/10.1177/0021998317729210.
\item Www.convergent.ca/products/compro-simulation-software. COMPRO simulation software 2014.
\item Fabris J, Mobuchon C, Zobeiry N, Lussier D, Fernlund G, Poursartip A. Introducing thermal history producibility assessment at conceptual design. Int. SAMPE Tech. Conf., vol. 2015- Janua, 2015.
\item Zobeiry N, Poursartip A. Theory-Guided Machine Learning for Process Simulation of Advanced Composites arXiv preprint arXiv:2103.16010 (2021).
\item Mirzaei S, Krishnan K, Al Kobtawy C, Roberts J, Palmer E. Heat transfer simulation and improvement of autoclave loading in composites manufacturing. Int J Adv Manuf Technol 2021;112. https://doi.org/10.1007/s00170-020-06573-3.
\item Zobeiry N, Stewart A, Poursartip A. Applications of Machine Learning for Process Modeling of Composites. 2020 Virtual SAMPE Conf., 2020.
\item Humfeld KD, Zobeiry N. Machine learning-based process simulation approach for real-time optimization and active control of composites autoclave processing. SAMPE Virtual Conf., vol. accepted, Long Beach, CA: 2021.
\item Zobeiry N, Humfeld KD. A physics-informed machine learning approach for solving heat transfer equation in advanced manufacturing and engineering applications. Eng Appl Artif Intell 2021;101:104232. https://doi.org/10.1016/j.engappai.2021.104232.
\item Kim M, Zobeiry N. Machine Learning for Reduced-order Modeling of Composites Processing. SAMPE Virtual Conf., vol. accepted, Long Beach, CA: 2021.
\item Zobeiry N, Reiner J, Vaziri R. Theory-guided machine learning for damage characterization of composites. Compos Struct 2020;246:112407. https://doi.org/10.1016/j.compstruct.2020.112407.
\item Chen CT, Gu GX. Machine learning for composite materials. MRS Commun 2019;9. https://doi.org/10.1557/mrc.2019.32.
\item Sacco C, Baz Radwan A, Anderson A, Harik R, Gregory E. Machine learning in composites manufacturing: A case study of Automated Fiber Placement inspection. Compos Struct 2020;250:112514. https://doi.org/10.1016/j.compstruct.2020.112514.
\item Califano A, Chandarana N, Grassia L, D’Amore A, Soutis C. Damage Detection in Composites By Artificial Neural Networks Trained By Using in Situ Distributed Strains. Appl Compos Mater 2020;27:657–71. https://doi.org/10.1007/s10443-020-09829-z.
\item Beck JV, Blackwell B, St Clair CR. Inverse Heat Conduction: Ill-Posed Problems. 1985.
\item Weber CF. Analysis and solution of the ill-posed inverse heat conduction problem. Int J Heat Mass Transf 1981;24. https://doi.org/10.1016/0017-9310(81)90144-7.
\item Van Ee D, Poursartip A. HexPly 8552 material properties database for use with COMPRO CCA and RAVEN, created for NCAMP 2009. 
\item Engl HW, Hanke M, Neubauer A. Regularization of Inverse Problems. Springer Science and Business Media; 1996.
\item Combal B, Baret F, Weiss M, Trubuil A, Macé D, Pragnère A, et al. Retrieval of canopy biophysical variables from bidirectional reflectance using prior information to solve the ill-posed inverse problem. Remote Sens Environ 2003;84:1–15. https://doi.org/10.1016/S0034-4257(02)00035-4.
\item Hochreiter S, Schmidhuber J. Long Short-Term Memory. Neural Comput 1997;9. https://doi.org/10.1162/neco.1997.9.8.1735.
\item HEXCEL. HexPly 8552 Epoxy matrix (180°C/365°F curing matrix) Product Data 2016. 
\item Hubert P, Johnston A, Poursartip A, Nelson K. Cure kinetics and viscosity models for Hexcel 8552 epoxy resin. Int. SAMPE Symp. Exhib., 2001.
\item Zobeiry N, Duffner C. Measuring the negative pressure during processing of advanced composites. Compos Struct 2018;203:11–7. https://doi.org/10.1016/j.compstruct.2018.06.123.
\item Www.convergent.ca/products/raven-simulation-software. RAVEN simulation software 2013.
\item Abadi M, Agarwal A, Barham P, Brevdo E, Chen Z, Citro C, et al. TensorFlow: Large-Scale Machine Learning on Heterogeneous Distributed Systems 2016.
\end{enumerate}
\end{document}